\documentclass[11pt, a4paper, logo, copyright, nonumbering]{googledeepmind}

\usepackage[authoryear, sort&compress, round]{natbib}
\title{\alg: VLM Guided Exploration of VLA Policies}

\reportnumber{001} % Leave blank if n/a

\newif\ifrevision
\revisionfalse
\author[*,1]{Bhavya Sukhija}
\author[1]{Oliver Groth}
\author[1]{Mohit Shridhar}
\author[1]{Tim Hertweck}
\author[1]{Michael Bloesch}
\author[1]{Markus Wulfmeier}
\author[1]{Abbas Abdolmaleki}
\author[1]{Martin Riedmiller}

\affil[*]{Work done as a Student Researcher in 2025}
\affil[1]{Google DeepMind}
\usepackage{amsthm} % better to load after ams math
\usepackage{mathtools,thmtools}

\usepackage{enumitem}
\usepackage{bm}
\usepackage{xspace}
\usepackage{nicefrac}
\usepackage{xfrac}
\usepackage{afterpage}

\definecolor{GoogleDarkRed}{RGB}{179, 20, 18}
\newcounter{todocounter}
\usepackage[capitalise]{cleveref}
\usepackage{subcaption}
\usepackage{dsfont}

\usepackage[textsize=footnotesize,textwidth=0.75in,
            color=cyan,bordercolor=white]{todonotes}
\newcommand{\ignore}[1]{}

\declaretheoremstyle[
	    spaceabove=\topsep, 
	    spacebelow=\topsep, 
	    bodyfont=\normalfont\itshape,
    ]{theorem}

\declaretheoremstyle[
	    spaceabove=\topsep, 
	    spacebelow=\topsep, 
	    bodyfont=\normalfont,
    ]{definition}

\declaretheorem[style=definition,numbered=no,name=Definition]{definition*}
\declaretheorem[style=definition,numbered=no,name=Remark]{remark*}
\declaretheorem[style=definition,numbered=no,name=Example]{example*}
\declaretheorem[style=definition,numbered=no,name=Question]{question*}

\usepackage{xifthen}

\newcommand{\likelihood}[2][]{%
    \ifthenelse{\isempty{#1}}
        {p\left(#2\right)}
        {p_{#1}\left(#2\right)}%
}

\newcommand{\normal}[3][]{%
    \ifthenelse{\isempty{#1}}
        {\mathcal{N}\left(#2, #3\right)}
        {\mathcal{N}\left(#1|#2, #3\right)}%
}

\newcommand{\alg}{\textsc{Eximo}\xspace}

\newcommand{\widefig}[2][\textwidth]{\makebox[\textwidth]{\includegraphics[width=#1]{#2}}}

\usepackage{listings}
\lstdefinestyle{promptstyle}{
  basicstyle=\ttfamily\scriptsize,
  breaklines=true,
  breakautoindent=false,
  breakindent=0pt,
  postbreak=\mbox{\textcolor{gray}{$\hookrightarrow$}\space},
  columns=fullflexible,
  keepspaces=true,
  frame=single,
  framesep=4pt,
  xleftmargin=6pt,
  xrightmargin=6pt,
  aboveskip=1em,
  belowskip=1em,
}

\RenewDocumentCommand{\H}{mo}{\mathrm{H}\IfValueTF{#2}{\left[#1\ \middle|\ #2\right]}{\parentheses*{#1}}}
\NewDocumentCommand{\Hsm}{mo}{\mathrm{H}\IfValueTF{#2}{[#1 \mid #2]}{\parentheses{#1}}}
\NewDocumentCommand{\I}{mmo}{\mathrm{I}\IfValueTF{#3}{\!\left(#1;#2\ \middle|\ #3\right)}{\parentheses*{#1; #2}}}
\NewDocumentCommand{\Ism}{mmo}{\mathrm{I}\IfValueTF{#3}{(#1;#2 \mid #3)}{\parentheses{#1; #2}}}

\newcommand{\E}{\mathbb{E}}

\def\m0{{\bm{0}}}

\def\va{{\bm{a}}}

\def\vs{{\bm{s}}}

\def\vx{{\bm{x}}}

\def\vpi{{\bm{\pi}}}

\def\setA{{\mathcal{A}}}

\def\setG{{\mathcal{G}}}

\def\setS{{\mathcal{S}}}
\def\setT{{\mathcal{T}}}

\usepackage{cancel}

\ifrevision
    \usepackage[draft]{changes}
    \definecolor{darkgreen}{rgb}{0,0.5,0}
    \setaddedmarkup{{\color{darkgreen}#1}}
    \setdeletedmarkup{{\color{red}#1}}
\else
    \usepackage[final]{changes}
\fi

\begin{abstract}
\emph{How to efficiently finetune robot policies to learn new tasks on the fly?}
State of the art robotic manipulation policies are based on behaviour cloning of large vision-language-action (VLA) models with billions of parameters on huge teleoperation datasets. While this simple approach has enabled significant advances for robotic manipulation, finetuning of VLA policies for learning new tasks still remains an open problem. In particular, collecting teleoperation datasets requires hundreds of hours of expensive human labour and the alternative, reinforcement learning (RL), can be notoriously sample-inefficient especially for long-horizon tasks. In addition, RL with VLAs imposes several challenges due to the model's size and architectural design. In this work, we propose \alg, an efficient algorithm for finetuning of VLA policies. \alg operates in three stages: \emph{explore}, \emph{imitate}, and \emph{optimize}.  During the explore phase,
\alg equips
the VLA with a vision language model (VLM) that acts as a planner. The VLM thinks and breaks down challenging long-horizon problems into shorter ones for the VLA. The VLM, together with the VLA, is used to collect an orchestrated dataset on new tasks. During the imitate phase, the VLA is finetuned with the orchestrated data. Finally, during the optimize stage, we use residual off-policy RL to further finetune the policy. In our experiments, we ablate all three stages of \alg and show that it outperforms existing approaches significantly in terms of sample-efficiency and final performance. 
\end{abstract}

\begin{document}

\maketitle

\section{Introduction}
% Defer the overview figure to the top of page 2 (reviewer request).
\afterpage{%
\begin{figure*}[t]
    \centering
    \resizebox{\textwidth}{!}{\input{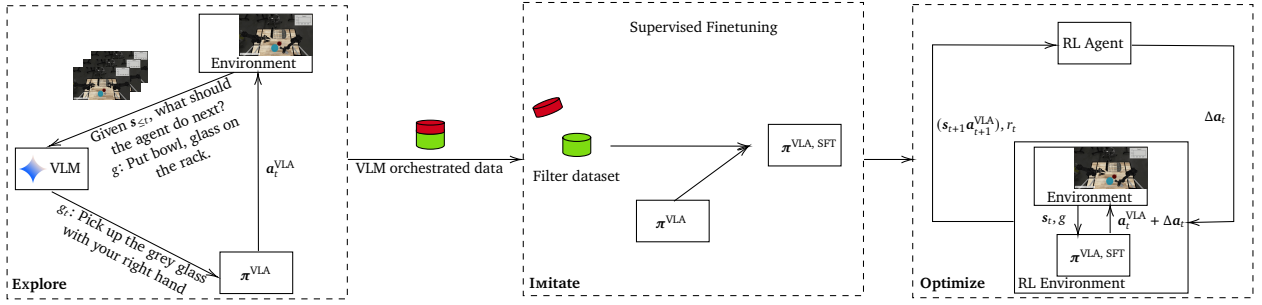}}
    \caption{\alg consists of three stages: Semantic exploration via VLM orchestration, where a VLM is used to command the VLA in natural language to accomplish a challenging task. The data from the VLM orchestration is used in the second phase for imitation learning, where the successful episodes are filtered and the VLA is finetuned via supervised finetuning on the filtered dataset. Finally, in the third stage, we use the finetuned VLA to perform residual reinforcement learning online.}
    \label{fig:overview}
\end{figure*}%
}
A crucial challenge in learning for robotics is to collect high quality datasets for training robot policies. Essentially, there are two approaches that have been widely applied for this purpose: (\emph{i}) reinforcement learning~\citep{sutton2018reinforcement}, and (\emph{ii}) large scale behavioural cloning (LBC)~\citep{brohan2022rt, black2410pi0, barreiros2025careful, gr15}. RL enables robotic agents to collect their own data and self-improve, and it has had great success in games and robotic locomotion tasks~\citep{mnih2013playing, miki2022learning} where high-fidelity and fast simulators are available. However, RL suffers from notoriously high sample complexity, especially for long-horizon problems where exploration is particularly challenging. This limits its direct application in the real world. To this end, large scale behavioural cloning is being widely adopted for robotic manipulation. Here teleoperators are used to collect high-quality and diverse motions with the robotic system and a policy is trained to match the motions from the teleoperators. This approach is simple and significantly more sample-efficient than RL. However, it suffers from generalization challenges, since the policy is trained only on the data distribution of the teleoperation and does not completely capture the diversity of the real-world. Furthermore, it is expensive since it requires significant amounts of human hours that are repeated for each new task and robotic system.  

In this work, we aim to bridge the gap between the two aforementioned strategies. Concretely, we consider the following setting: we are given a pretrained, language-conditioned vision-language-action (VLA) policy, itself trained via LBC, that reliably performs a repertoire of \emph{atomic} skills (e.g., pick and place) within its training distribution, but struggles on \emph{new}, long-horizon, compositional, or reasoning-heavy tasks that require chaining these skills together or interpreting goals that lie outside the training distribution (e.g., ``put the fruit the monkey likes to eat in the bowl''). Our objective is to adapt the VLA to such tasks \emph{sample-efficiently} and \emph{without additional teleoperation}, where each task is specified only through a natural-language goal and a success detector. Instead of relying on teleoperators, we leverage the world-knowledge embedded in state-of-the-art vision-language foundation models (VLMs) to orchestrate the VLA and collect data on these more complex, long-horizon tasks. Next, we distill the knowledge of the VLM orchestrator by supervised fine-tuning of the initial policy with the newly collected data. This enables us to teach the policy new skills without requiring expensive teleoperation hours. Finally, we further finetune the initial policy using online reinforcement learning. After the fine-tuning phase, the policy is already capable of achieving non-trivial performance, making the exploration for online RL tractable, i.e., more sample-efficient. In summary, we use a three-step approach; (\emph{i}) data-collection with VLM orchestration, (\emph{ii}) supervised-finetuning on the collected data, (\emph{iii}) further finetuning of the policy via online RL. The algorithm, \alg, is depicted in \cref{fig:overview}.

We evaluate \alg on \emph{twenty-two} different manipulation tasks in simulation that require reasoning about the object to manipulate and chaining skills together to achieve success. In our experiments, we show that VLM orchestration results in significant sample-efficiency gains, and that \alg outperforms both the SOTA base VLA---even when the latter is given additional RL finetuning---and the base VLA combined with VLM orchestration.

%Furthermore, we also evaluate steps (\emph{i}) and (\emph{ii}) on the aloha platform~\citep{aldaco2024aloha}.

\section{Related Work}
\paragraph{Foundation Models as High-level Planners}
\looseness=-1
Several works leverage LLMs as high-level planners for robotic tasks~\citep{huang2022language,  ahn2022can, wu2023tidybot, bhat2024grounding, mon2025embodied}.
In particular, \citet{huang2022language} use LLMs as a receding horizon planner for long-horizon problems. The LLM generated actions are then grounded to permissible actions using a masked LLM. Generally, grounding is a crucial challenge for LLM planners.
\citet{huang2023instruct2act, singh2023progprompt, wu2023tidybot, mon2025embodied} provide robot APIs in the prompt to the LLMs for grounding,
 whereas \citet{ahn2022can} pass a finite set of skills along with their probability of success to the LLM planner as context for grounding.  \citet{bhat2024grounding} propose using two LLMs, one for generating action plans in natural language and another LLM for grounding the actions to robot API calls. Crucially, recent advances in VLA policies, e.g.,  Gemini Robotics On-Device (\href{https://deepmind.google/discover/blog/gemini-robotics-on-device-brings-ai-to-local-robotic-devices/}{GROD}) and Gemini Robotics~\citep{gr15}, enable directly interacting with the robots in natural language, alleviating the grounding challenges associated with LLM planners. \citet{shi2025hi} jointly train a VLM and VLA, where the former is used to decompose high-level natural language instructions into executable low-level ones. The low-level instructions are then passed to the VLA which generates the actions. In this work, we focus on the finetuning of VLAs using SOTA VLM models.
Moreover, we leverage a natural language instructable VLA policy and study the problem of sample-efficient finetuning of the policy on challenging robotic tasks. 

\paragraph{Foundation Models for Exploration}
\looseness=-1
\citet{dalal2024plan} use LLMs to decompose challenging long-horizon problems into shorter-horizon subtasks and train an RL agent to execute the LLM generated plan. They show that leveraging the planner provides significant sample-efficiency gains.  However, they also focus on the grounding of LLM generated plans and train a vision policy from scratch via RL. Instead, we study the finetuning of SOTA VLA policies which are instructed by the VLM directly in natural language. We show that VLM orchestrated exploration yields significant sample efficiency gains in the finetuning of the VLA without requiring any additional grounding. 

Foundation models are also used to guide intrinsic exploration of agents. In particular, \citet{tam2022semantic} use representations present in pretrained VLMs to quantify novelty for intrinsic exploration. \citet{sancaktar2025sensei} use VLMs to measure the interestingness of states for exploration and learn an intrinsic reward function via preference learning from the VLM feedback. \citet{du2023guiding} study goal-conditioned RL and leverage causal LLMs to propose goals for exploration and masked LLMs to reward accomplishing the proposed goals. Similarly, \citet{zhang2023bootstrap} use LLMs to explore a sequence of skills from a finite skill set.

\paragraph{RL Finetuning of VLAs}
RL enables agents to self-improve by directly interacting with the environment. However, SOTA VLA policies are typically large diffusion and/or transformer models~\citep{brohan2023rt1roboticstransformerrealworld, barreiros2025careful, black2410pi0} that are trained to predict action chunks~\citep{brohan2023rt1roboticstransformerrealworld}. This makes training these models via RL extremely challenging. To this end, \citet{mark2024policy} learn a Q-function to optimize actions instead of training the policy, whereas \citet{ankile2025imitation, ankile2025residual} learn a residual policy to correct actions of the VLA. We take a similar approach as \citet{ankile2025residual} and learn a residual policy via off-policy RL. However, we additionally improve sample-efficiency of our method by integrating VLM guided exploration.

\section{Method}
In the following, we describe \textbf{\textsc{Ex}}plore \textbf{\textsc{Im}}itate and \textbf{\textsc{O}}ptimize (\textbf{\alg}), our three-step training pipeline for sample-efficient robot learning. 
\subsection{\textsc{Ex}plore: VLM-Guided Data Collection}
\begin{figure*}[ht]
    \centering
    \resizebox{\textwidth}{!}{\input{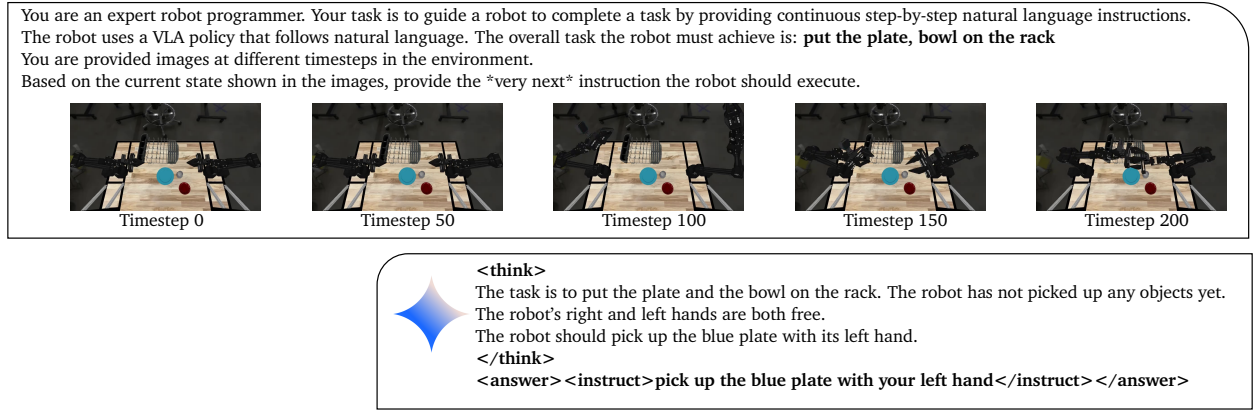}}
    \caption{Example interaction of the VLM during the explore phase of \alg. The VLM is given a sequence of images from the environment along with the task description in the prompt. The VLM analyzes the provided information within a <think> </think> block and provides an instruction in the <answer></answer> block for the VLA to execute. }
    \label{fig:example_interaction}
\end{figure*}
We assume access to an initial goal-conditioned manipulation policy that is capable of performing fundamental skills such as pick and place. Moreover, we use 
the 3B variant of Gemini Robotics On-Device~\citep{grod} as our initial policy. GROD is a vision-language-action (VLA) model based on the PaliGemma~\citep{beyer2024paligemma} VLM backbone and diffusion policy head. GROD is trained on manipulation tasks with teleoperated data for the Aloha robot in both real and simulation~\citep{aldaco2024aloha}. Given the state measurement $\vs \in \setS$ and a goal in \emph{natural language} $g \in \setG \subseteq \setT^L$\footnote{Here $\setT$ represents the set of tokens in the vocabulary of the model and $\setT^L$ a sequence of tokens.}, e.g., ``pick up the yellow banana'' the VLA returns an action $\va \in \setA$ attempting to accomplish the goal. That is $\va \sim \vpi^{VLA}(\cdot|\vs, g)$. However, while the model has high success rates in tasks that lie within the training distribution, e.g., ``put the banana in the bowl'', its performance drops in goals and states that lie outside of training data, e.g., ``put the fruit the monkey likes to eat in the bowl'' or ``put the banana and lime in the bowl''. In general, collecting data for all plausible tasks via teleoperation is intractable. However, since the model is capable of performing fundamental skills, we combine it with a state-of-the-art VLM, in particular Gemini~\citep{team2023gemini}, which decomposes the goal $g$ into intermediary goals/skills that the VLA is capable of executing. Moreover, the VLM acts as an orchestration policy which, at time $t$, given the history of states $\vs_{\leq t}$ and the natural goal $g$ provides tractable goals $g_t$ for the VLA to reach, i.e., $g_t \sim \vpi^{VLM}(\cdot|\vs_{\leq t}, g)$ (see~\cref{fig:example_interaction}). The VLA is then conditioned on the goal provided by the VLM. We integrate the VLM and VLA interaction in a closed-loop manner, allowing the VLM to change and adapt the goals provided to the VLA as and when needed. 

This approach provides a natural separation between fundamental physical skills that require high quality and domain specific teleoperated datasets with which VLAs are trained and semantic understanding of the underlying task for which SOTA VLMs can be leveraged off-the-shelf.  
We rollout the VLM-orchestrated policy to collect data for new unseen tasks and store the rollouts that successfully accomplished the task in a data buffer, which we use for supervised fine-tuning. Whether a rollout was successful is determined by the environment's ground-truth success detector: at the end of each episode we query the task-success signal and only add episodes marked successful to the buffer, thereby filtering out unsuccessful orchestration attempts.\footnote{We assume access to a ground-truth success detector, as is standard in simulation. Replacing it with a VLM-based success detector is a promising direction discussed in the \nameref{sec:future} section.}

\subsection{\textsc{Im}itate: Supervised Finetuning on VLM Orchestrated Data}
We collect a dataset consisting of states, goals, and the corresponding action, i.e., $(\vs, g, \va)$ and finetune the VLA to predict the action $\va$  given $(\vs, g)$\footnote{In practice, instead of learning to predict the immediate action $\va$, we train the VLA to predict a chunk of next actions, i.e., $\va_{0:K-1}$ where $K$ is the action chunk size. See~\cite{zhao2023learning} for more detail.}. Crucially, here instead of keeping the VLM orchestrated-goal, we train the VLA to directly predict the actions conditioned on the actual task goal $g$. Through this we distill the knowledge of the VLM orchestrated policy back into the VLA. Concretely, finetuning here refers to continuing to train the pretrained GROD policy on the filtered orchestrated dataset using the same behaviour-cloning objective with which the VLA was originally trained, i.e., we update the model weights to predict the action chunk $\va_{0:K-1}$ from $(\vs, g)$ via gradient descent. We report the exact VLM orchestration prompt in \cref{fig:orchestration_prompt} and the full task suite in \cref{tab:tasks}.
This approach has several benefits: first, it enables
efficient real-time control on the system, since the VLM latency issues are circumvented by distilling the knowledge directly into the much smaller VLA; furthermore, it also significantly minimizes VLM calls, as the VLM is not queried any further after data collection. Finally, it simplifies the downstream RL pipeline because the RL agent now only interacts with the VLA. This also answers a natural question---why not simply keep the VLM orchestrator at evaluation time? Beyond removing the VLM's latency, distillation even improves task performance over orchestrating at evaluation time, as we show in \cref{fig:post_sft} (GROD + SFT vs.\ GROD + VLM-Orchestration).

\subsection{\textsc{O}ptimize: Finetuning via Online RL}
In the final stage, we further finetune the policy via online RL. RL finetuning of VLAs is particularly challenging due to their large size and implicit policy distribution induced from the diffusion process. Therefore, we instead use a residual policy similar to \citet{ankile2025residual}. The policy outputs a residual action $\Delta \va$. This residual action is then combined with the action returned by the VLA and then executed on the system, i.e., $\va = \va^{\text{VLA}} + \Delta \va$. The residual policy is conditioned on the current state, goal, and the VLA action, that is, $\Delta \va \sim \vpi^{\text{ref}}(\cdot|\vs, g, \va^{\text{VLA}})$. Accordingly, the residual MDP has the following state $\vx = (\vs, \va^{\text{VLA}})$ and given the action $\Delta \va$, the next state $\vx'= (\vs', \va^{'\text{VLA}})$ is obtained via 
the following transition dynamics
\begin{align*}
    \vs' &\sim T(\cdot|\vs, \va^{\text{VLA}} + \Delta \va) \\
    \va^{'\text{VLA}} &\sim \vpi^{VLA}(\cdot|\vs', g) \\
\end{align*}
Here $T$ represents the transition kernel of the underlying system. 

We train the residual policy to optimize the underlying probability of success
\begin{equation}
\max_{\vpi^{\text{ref}} \in \Pi}    \underbrace{\E_{\vpi^{\text{ref}}, \vs_0 \sim \rho}\left[\sum^{\infty}_{t=0}\gamma^t \mathbb{1}_{\vs_t \in \text{Success}(g)}\right]}_{J(\vpi^{\text{ref}}, \rho)} \label{eq:sparse reward problem}
\end{equation}
The set $\text{Success}(g) \subset \setS$ denotes the states for which the goal $g$ is accomplished. For most goals, this set is much smaller than $\setS$ and therefore $\mathbb{1}_{\vs_t \in \text{Success}(g)}$ represents a sparse reward, making the problem in \cref{eq:sparse reward problem} extremely challenging for RL. However, if the VLA policy is able to accomplish non-trivial success rates on the task, this improves the residual policy learning and exploration. We use the MPO algorithm~\citep{abdolmaleki2018maximum} for the RL finetuning of the residual controller. 
\section{Experiment}
\looseness=-1 We empirically validate our approach across several manipulation tasks for the ALOHA~\citep{aldaco2024aloha} robotic platform in simulation. In particular, we ablate each step of \alg and investigate the following questions: (\emph{i}) Does VLM orchestration with GROD yield better exploration and efficient data collection? (\emph{ii}) Does finetuning of GROD using the orchestrated data improve the model's performance? (\emph{iii}) Does online RL with the finetuned GROD result in more sample-efficient online learning?
\begin{figure*}[p]
    \centering
    % Left: the three barplots, cropped to identical size (per-plot legend and
    % title trimmed off) so all bars are the same size. Right: the shared
    % Baselines + Task Names legend, shown once (cropped from the EpisodeLength
    % asset). trim order = {left bottom right top}, values in bp.
    \begin{minipage}[c]{0.58\textwidth}
        \centering
        \includegraphics[width=\linewidth,trim={0 0 342 22},clip]{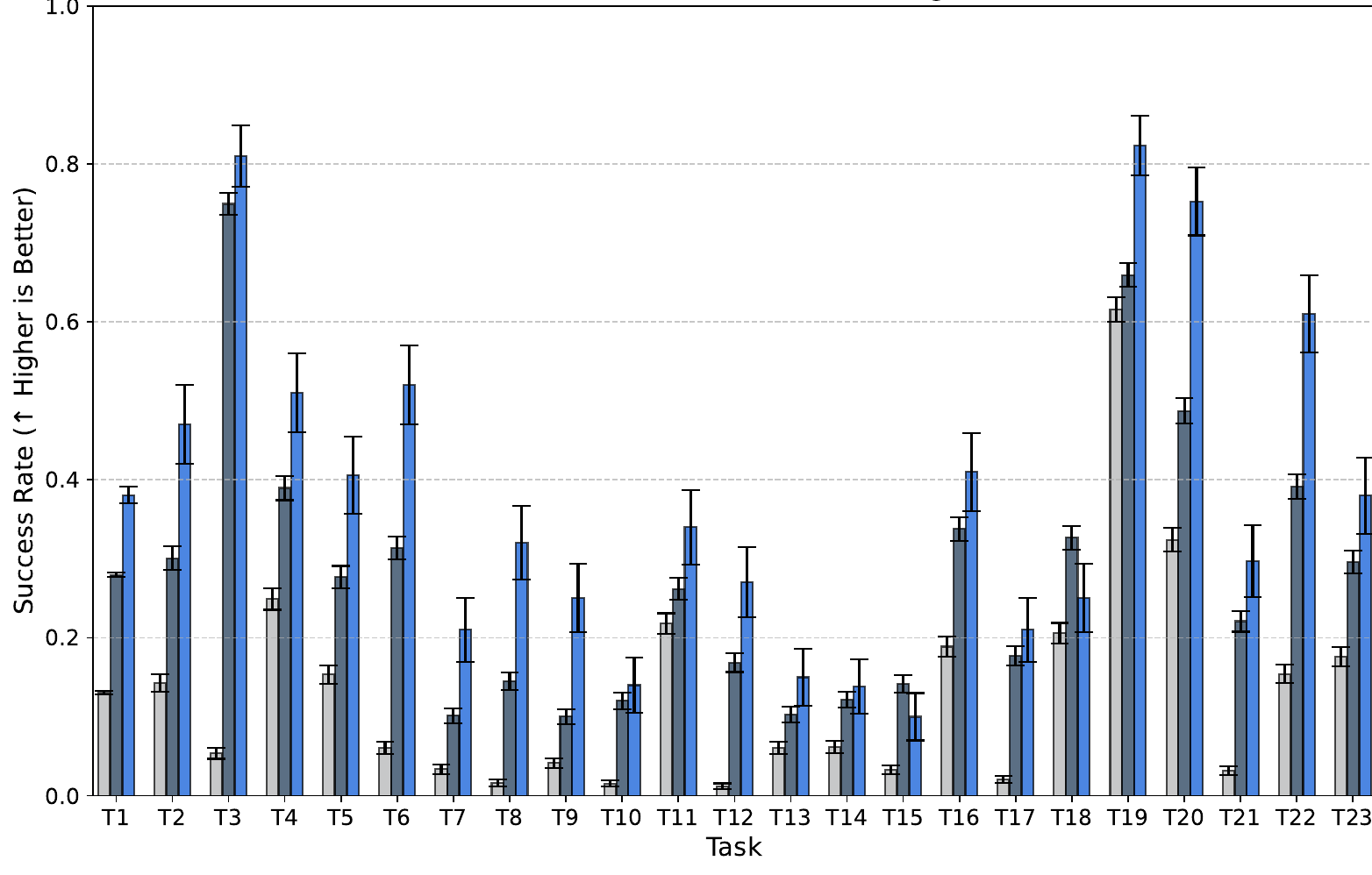}\\[1pt]
        \includegraphics[width=\linewidth,trim={0 0 4 5},clip]{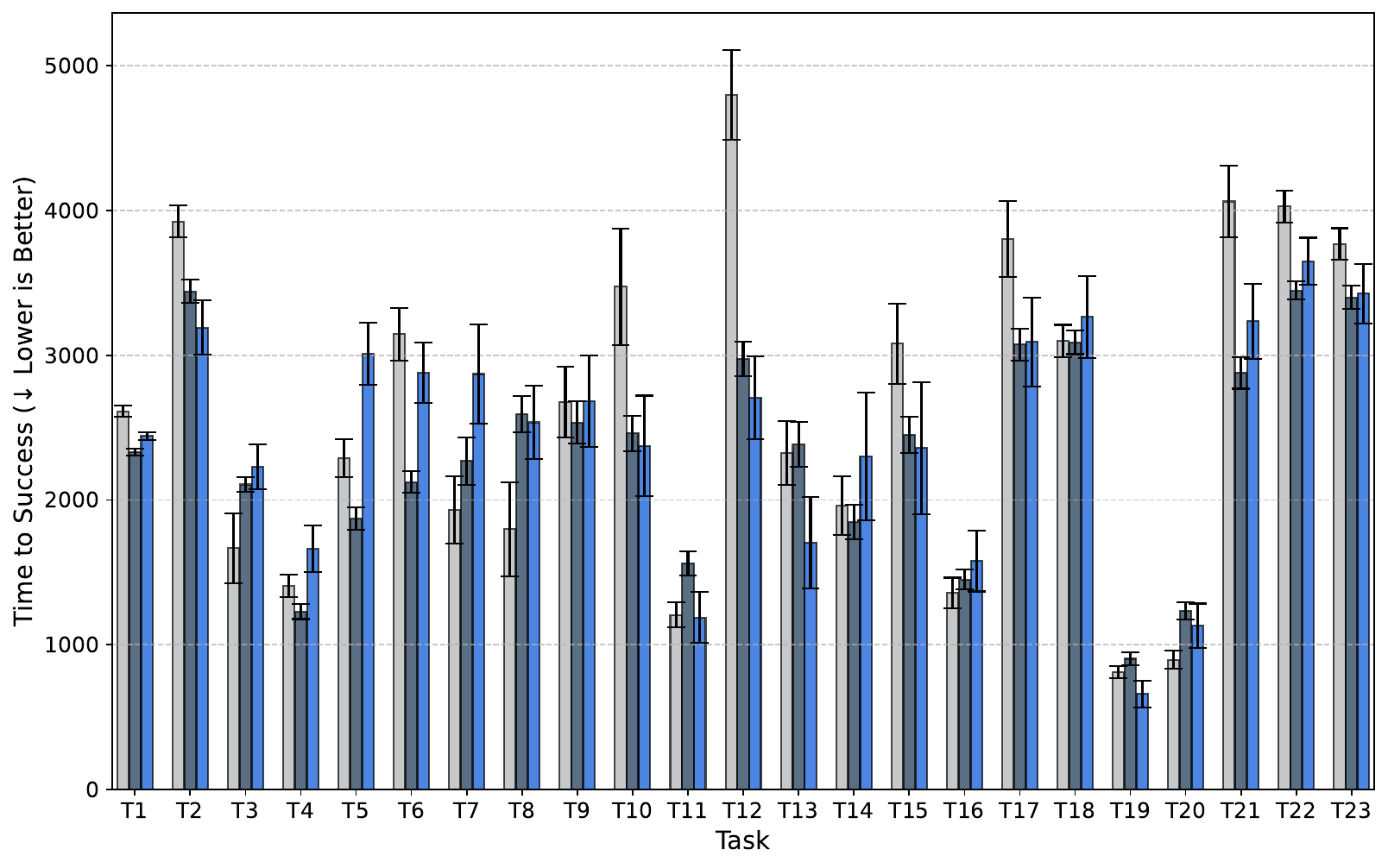}\\[1pt]
        \includegraphics[width=\linewidth,trim={0 0 334 22},clip]{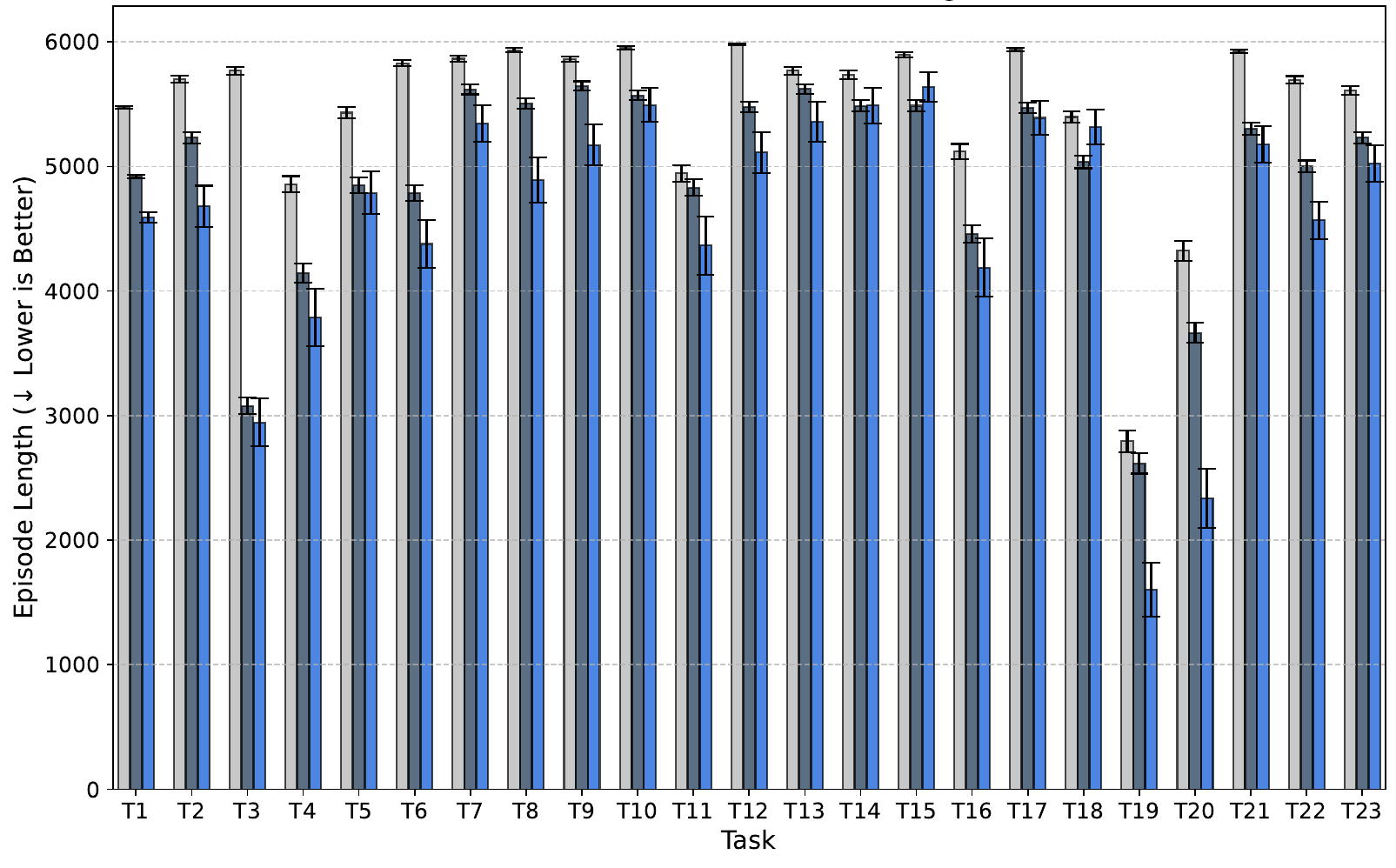}
    \end{minipage}\hfill
    \begin{minipage}[c]{0.40\textwidth}
        \centering
        \includegraphics[width=\linewidth,trim={768 0 0 0},clip]{assets/vlm_as_policy/post_sft_comparison_EpisodeLength_after_sft.pdf}
    \end{minipage}
    \caption{Success rate (top), time to success (middle), and episode length (bottom) of VLM orchestrated GROD, GROD with no orchestration, and GROD finetuned on the data collected from VLM orchestration. The baselines and the per-task legend (shared across all three plots) are shown on the right. Across the tasks, we observe that VLM orchestration achieves higher success rate than the model without orchestration. This illustrates the benefits of semantically guided exploration from the VLM. Furthermore, performing filtered SFT on the VLM orchestrated data gives additional performance gains in the success rate across the task, showcasing the advantages of distilling the VLM orchestrated trajectories into the VLA. The VLM orchestrated VLA also has a lower episode length and is therefore more data efficient. Note that time to success is only meaningful for tasks with non-zero success rate.}
        \label{fig:post_sft}
\end{figure*}
\subsection{Does VLM orchestration with GROD yield better exploration?} 
We compare the performance of the base VLA (no orchestration) to the VLA with VLM orchestration. As performance metrics, we consider the model's success rate and time-to-success. The results are reported in \cref{fig:post_sft}. 
We evaluate both methods across twenty two different tasks for \emph{1000} episodes. Each episode is terminated if the policy successfully solves the task. %Therefore, we also report the length of an episode as a measure for sample-efficiency (\cref{fig:episodelength}). All the results are reported with the mean performance and standard error. 

Despite GROD being a SOTA VLA, in \cref{fig:post_sft} we see that VLM orchestration significantly increases the success rate of GROD while maintaining similar time to success. The gain in success rate is particularly observable in long horizon tasks such as PlateBowlOnRack, that require chaining skills of the base VLA together. Furthermore, the VLM orchestration also improves the performance of the base VLA on reasoning tasks (e.g., BananaInBowl-Reasoning),  which require the agent to reason about the objects in the scene.
This underlines the benefits of leveraging SOTA foundation models for semantic exploration and VLA for accomplishing individual base skills.  Moreover, as shown in \cref{fig:post_sft} (bottom), the VLM orchestrated policy has much smaller episode lengths and accordingly is more sample-efficient in data collection.

In summary, via VLM orchestration, we can collect higher quality episodes, i.e., those with higher success rate, more data efficiently (since each episode is effectively shorter with VLM orchestration). In the following, we investigate whether this data can be leveraged for efficient finetuning of the base VLA.

\subsection{Does finetuning of GROD using the orchestrated data improve performance?}
Next, we compare the finetuned GROD agent with the VLM orchestrated and non-orchestrated agent. The results are presented in \cref{fig:post_sft}. After finetuning of the base model on the orchestrated data, we improve the performance of the base VLA drastically. Moreover, the finetuned VLA also outperforms the orchestrated agent, showcasing its ability to learn novel tasks/behaviours from the orchestrated data. We attribute these gains to two factors: (\emph{i}) the SFT dataset contains only the successful, filtered orchestration episodes, providing high-quality supervision, and (\emph{ii}) distillation compiles the multi-step, VLM-guided behaviour into a single policy conditioned directly on the task goal, so that the resulting VLA no longer depends on online VLM queries at evaluation time. From this, we conclude that VLMs can control and guide the exploration of VLAs, and that distilling the newly acquired skills back into the base VLA model yields a significant performance boost. This result paves the way for a new paradigm in VLA training, where VLAs trained on basic robotic skills can be used in conjunction with VLMs to acquire new skills without any additional teleoperation hours.

\subsection{Is online RL with the finetuned GROD more sample-efficient?}
\begin{figure*}[!ht]
\centering
% Two equal-width curves within the side margins; the shared Baselines legend
% lives inside the right (time-to-success) plot.
\includegraphics[width=0.49\textwidth]{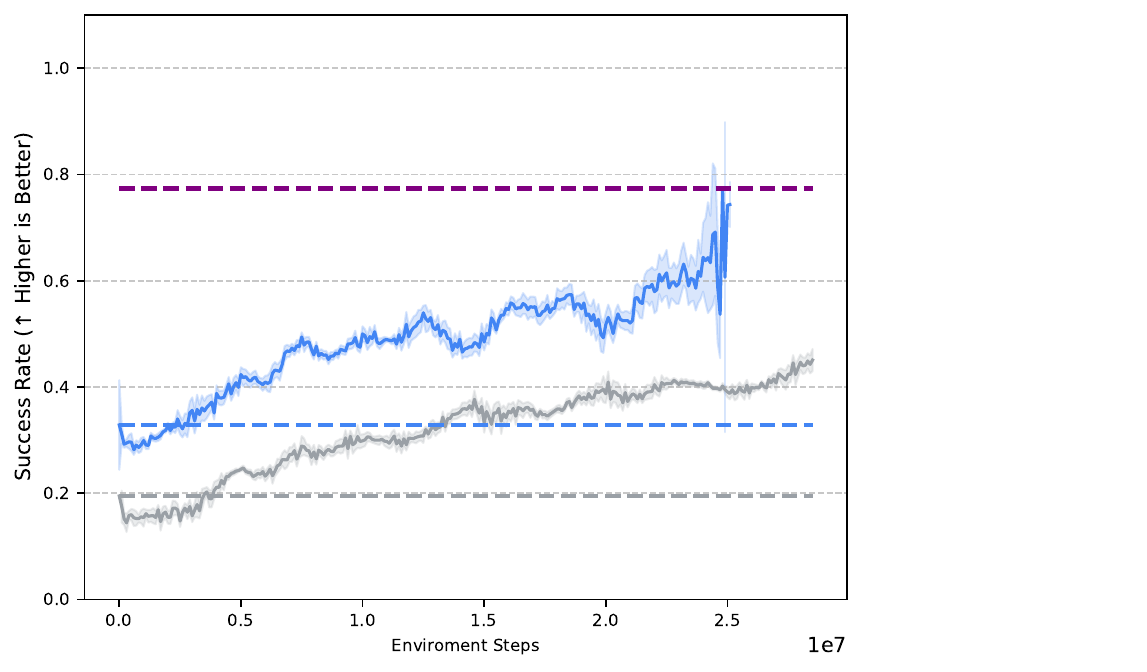}%
\hfill
\includegraphics[width=0.49\textwidth]{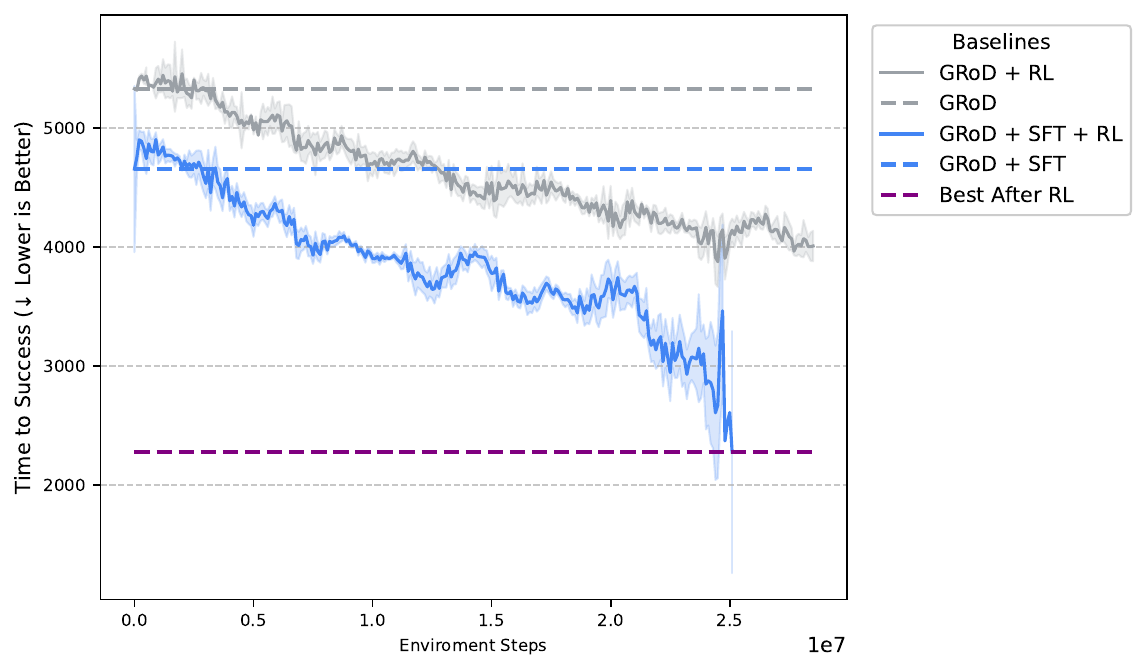}
    \caption{Online RL performance of GROD + SFT and the base GROD model averaged across twenty tasks. We run the base GROD model for more timesteps to compensate for the additional data collected via VLM orchestration for the SFT.  
    Due to the finetuning on VLM orchestrated data, GROD + SFT starts with a higher success rate. The model also converges to higher success rates and the base GROD model does not achieve the same performance, despite running online RL for significantly more environment steps. We observe the same behaviour for time-to-success. The performance is averaged across five seeds and we report the mean with two standard errors. Note that time to success is only meaningful for tasks with non-zero success rate. }
        \label{fig:rl_curve}
\end{figure*}
\begin{figure*}[!ht]
 \begin{subfigure}[b]{\textwidth}
        \centering
        \widefig{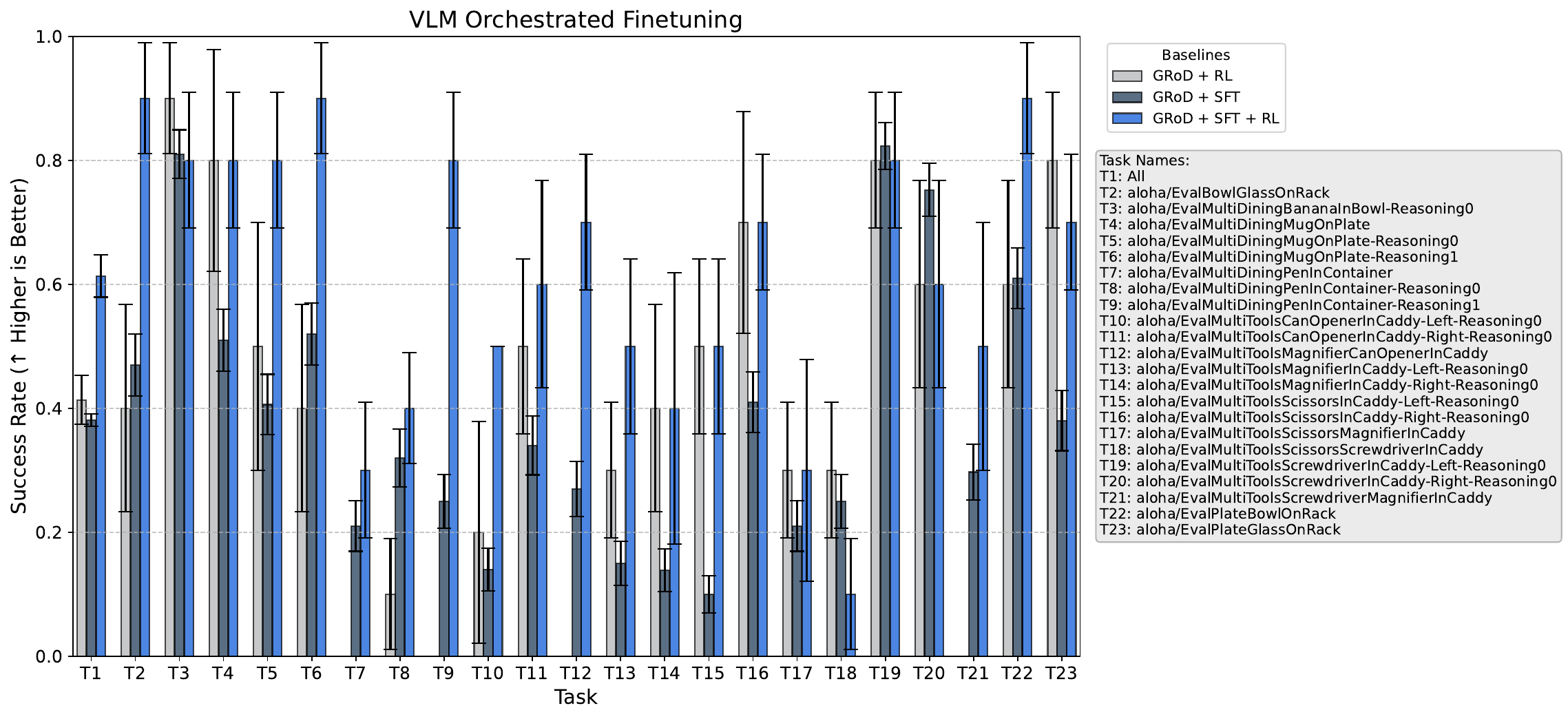}
       % \caption{Caption for the first subfigure}
      %  \label{fig:sub1}
    \end{subfigure}
    \hfill % Add some horizontal space between subfigures
    % Second subfigure
    \begin{subfigure}[b]{\textwidth}
       % \centering
        \widefig[0.82\textwidth]{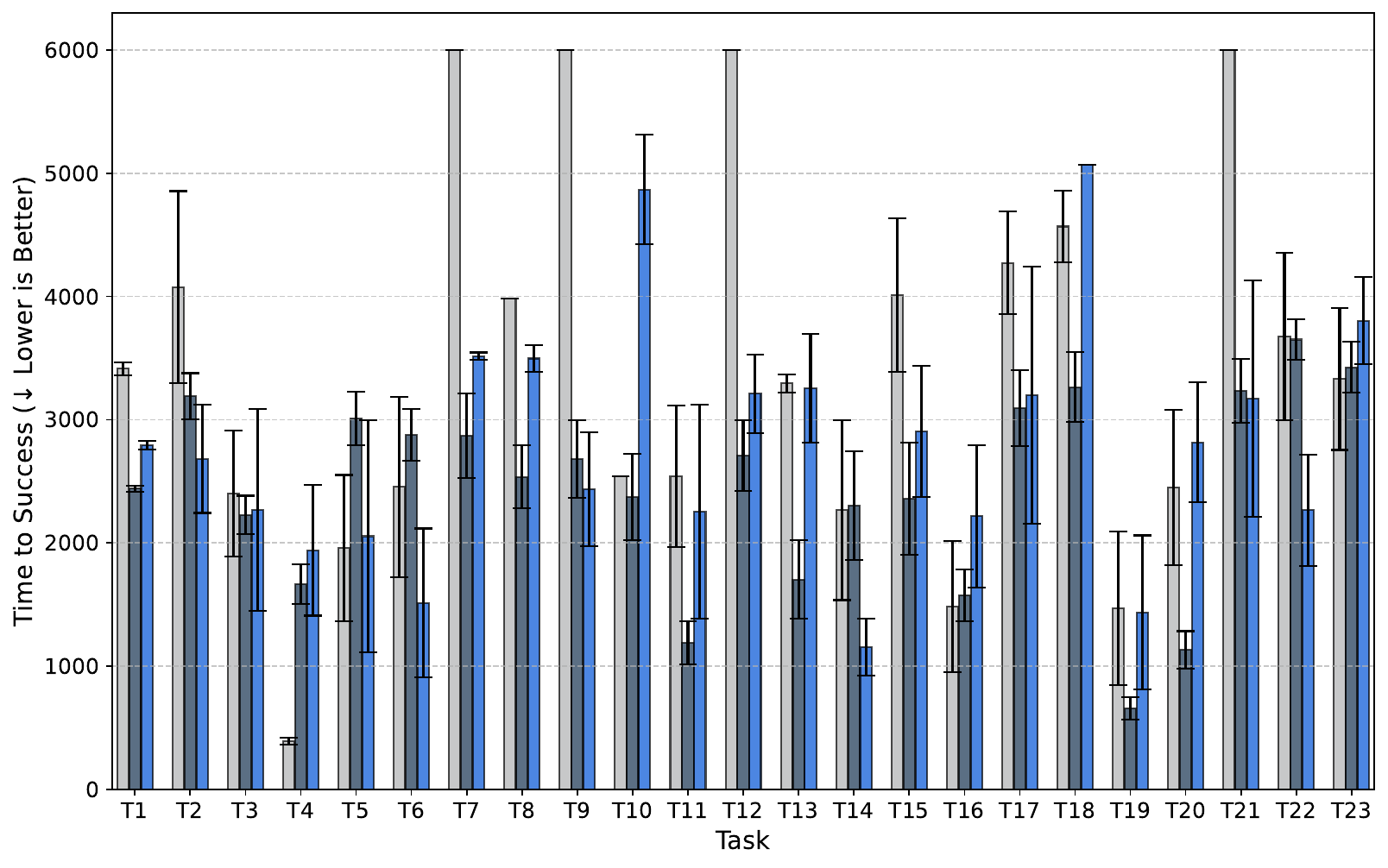}
    %    \caption{Caption for the second subfigure}
       % \label{fig:sub2}
    \end{subfigure}
    \caption{Success rate and time to success post RL finetuning. We compare the performance of the VLA after SFT on VLM orchestrated data, after SFT + RL and the base GROD model after RL. Across all tasks, we observe that the model after SFT + RL outperforms the other two baselines. Note that time to success is only meaningful for tasks with non-zero success rate.}
        \label{fig:results_post_rl}
\end{figure*}
We compare the performance after RL of the finetuned GROD VLA, GROD + SFT, with the base GROD model. Moreover, \citet{ankile2025residual} propose RL finetuning of BC policies using off-policy residual RL. We adopt the same approach for finetuning GROD.
To have a fair comparison between the SFT GROD model and the base model, we run RL on the base VLA for more episodes to compensate for the additional data collection performed during the Explore stage of \alg. We report the average success rate and time-to-success across all twenty tasks in \cref{fig:rl_curve}. From the figure, we conclude that
online RL improves the performance of both models. However, GROD + SFT starts at a higher success rate than the base model and also obtains higher performance at convergence compared to the base GROD model. Moreover, even though we run the base GROD model for more environment steps, it is not able to obtain the same performance as GROD + SFT. This is because GROD + SFT is trained on the higher quality data collected via VLM orchestration during the explore phase of \alg. The same behaviour is present for time-to-success. 

In \cref{fig:results_post_rl} we report the final performance after SFT and after SFT + RL across all tasks. We see that online RL further boosts the success rate of the agent consistently across the tasks. This demonstrates the benefits of the optimize phase of \alg.  Furthermore, on several tasks SFT on only VLM orchestrated data outperforms the RL finetuned base VLA, despite the latter collecting significantly more data. This illustrates the benefits of VLM orchestrated data collection and distillation into the VLA model.

% \begin{darkredtext}
% \paragraph{Hardware Experiments} We evaluate the explore and imitate stage of \alg on the Aloha platform. Here the goal is to show the following:
% \begin{enumerate}
%     \item VLM Orchestration on top of the GROD model improves success rates and provides more efficient data collection on HW for new/unseen tasks.
%     \item Finetuning the GROD model on the orchestrated data improves the performance of the agent on the new/unseen tasks.
% \end{enumerate}
% We will evaluate 3-5 tasks and collect 100 episodes for each task with this setup. In total corresponding to 300-500 episodes worth of data used for the finetuning.  
% \end{darkredtext}

\section{Conclusion}
In this work, we showcase the benefits of combining the general knowledge embedded in VLMs with the sensory-motor skills of VLAs. We use the VLM to explore efficiently together with the VLA by breaking down complex tasks into intermediary steps that the VLA can execute. We show that this approach yields a higher success rate than using the base VLA alone. Furthermore, we finetune the VLA to distill the trajectories collected with VLM orchestration, thereby improving the performance of the VLA on new tasks. Next, we further improve the policy via off-policy residual RL. Across several tasks on the Aloha benchmark suite, we show that our approach significantly outperforms the baselines in terms of both sample-efficiency and performance. 

\section{Future Work}\label{sec:future}
 In the current setup, we assume access to ground truth success detectors. However, in principle, VLMs can also be used to detect whether a task has been completed. Future work will focus on leveraging VLMs not only as orchestrators but also as success detectors for learning. Furthermore, we also plan to leverage the VLMs for resetting the environment by orchestrating the agent to ``undo the task''. For reversible tasks, this would enable a fully autonomous learning loop where the VLM is used for both reward modelling, orchestration, and resets.  

 More broadly, the exploration and imitation phases of \alg can be viewed as a post-training procedure for VLAs based on self-distillation~\citep{lu2025onpolicydistillation, hubotter2026reinforcement}. Our approach distills additional contextual information derived from task breakdowns produced by the VLM back into the VLA, yielding a stronger standalone policy without requiring orchestration at deployment time. In this work, we restrict ourselves to an off-policy setting, in which trajectories collected during exploration are reused for supervised fine-tuning during imitation. A promising direction for future work is to extend this framework to on-policy distillation methods~\citep{hubotter2026reinforcement, shenfeld2026self}.

 % To this end, we will use the VLM orchestrator in conjunction with the VLM as success detector. Here the data in the imitate phase will be filtered using the VLM feedback instead of the ground truth success signal. We will evaluate this approach in simulation on five tasks. The following will be investigated
% \begin{enumerate}
%     \item Finetuning the GROD model on the orchestrated data with VLM as success deterctor improves the performance of the agent on the new/unseen tasks.
% \end{enumerate}
% We will evaluate 5 tasks and collect 100 episodes for each task with this setup. In total corresponding to 500 episodes worth of data used for the finetuning.
% Another section. See \cite{silver2016mastering} for reference.

\bibliography{main}
\clearpage

\appendix
\section{Additional Experiments}

\begin{figure}[!ht]
    \centering
    \widefig{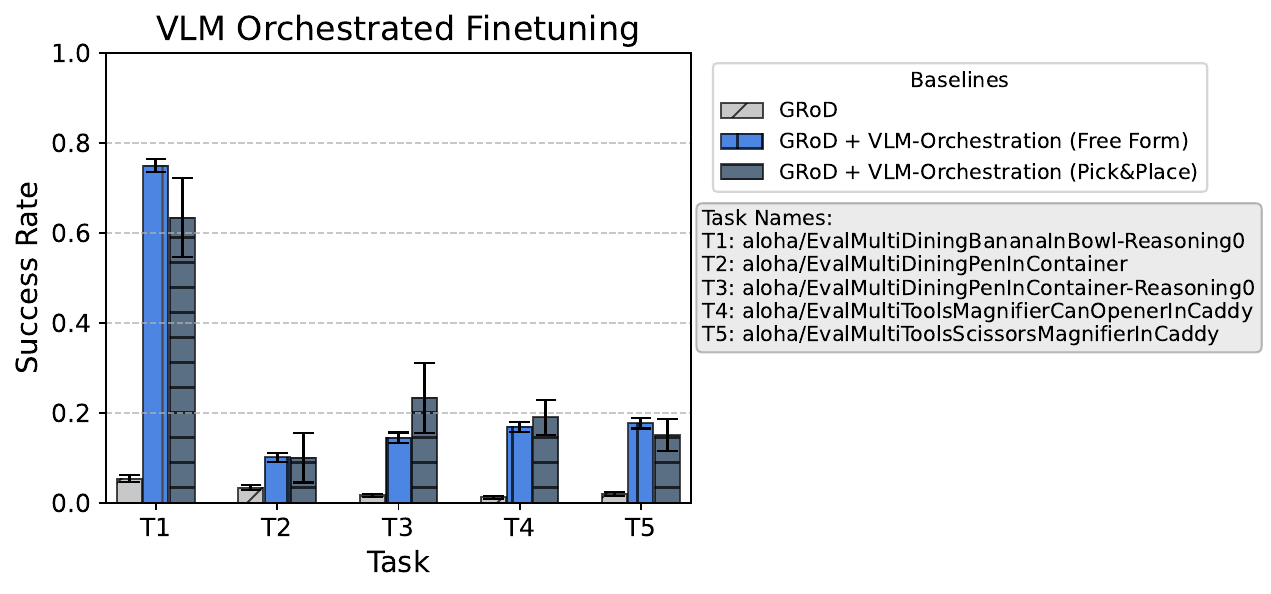}
    \caption{Performance of VLM orchestrated GROD with free form orchestration and only Pick\&Place commands. The former commands the model in natural language, whereas the latter only provides pick and place commands. We find that GROD benefits from the VLM orchestrators and is robust to the types of orchestration, performing on par for both free form and Pick\&Place orchestration.}
    \label{fig:instruction_vlm_comparison}
\end{figure}

\paragraph{Do free form instructions from VLM affect performance?}
A key advantage of using GROD as the base VLA is that it can be commanded in free form natural language by the VLM and does not require any additional grounding\footnote{In our experiments, the VLM gives natural language commands such as ``move the yellow banana further into the light blue bowl with your left hand''}. To illustrate these capabilities of GROD, we compare it with a grounded baseline, where the VLM is restricted to only giving pick and place commands, i.e., pick up the glass\footnote{The free form version can instead describe the object, provide additional specifications, or use different wordings, e.g., pick up the grey glass with your left hand or drop the grey glass.}.
In \cref{fig:instruction_vlm_comparison} we compare the free form instruction VLM with the grounded one. We see that both the free form and pick\&place instructions lead to comparable performance across the five tasks considered. This aligns with our intuition of the GROD VLA, which is trained to follow natural language instructions and therefore can be commanded by the VLM directly without any additional grounding.

\begin{figure}[!ht]
    \centering
    \widefig{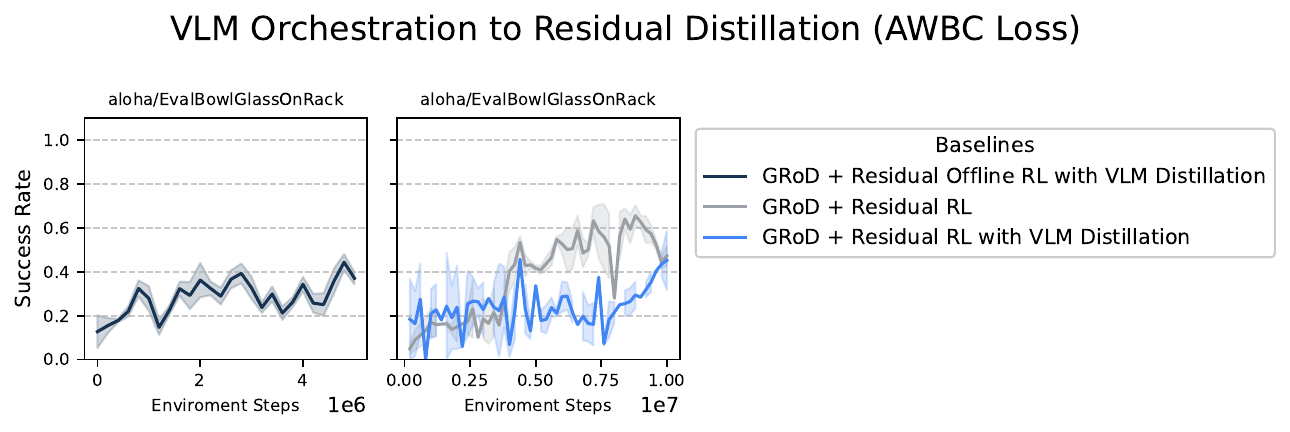}
    \caption{Performance of VLM to residual policy distillation using AWBC for offline RL and residual RL during the online phase. Left: Performance during offline RL as a function of the offline dataset size. Right: Performance during the online RL phase. We observe that while the residual policy learns during the offline RL phase, it fails to benefit from it during the online RL one. This is most likely due to the distribution shift between the VLM orchestrated data (offline) and online rollouts. These results are obtained across three seeds and the mean performance with standard deviation is reported.}
    \label{fig:AWBCLoss Distillation}
\end{figure}

\paragraph{Distilling VLM Orchestration to the Residual Policy}
We investigate distilling the VLM orchestrator into the residual policy instead of the GROD VLA. In order to achieve this, we first collect the orchestrated dataset using the base VLA and store the tuple: $(\vx, \Delta \va, \vx', r)$, where $\vx = (\vs, \va^{VLA})$, $\Delta \va = \va^{VLA-VLM} - \va^{VLA}$, and $\vx' = (\vs', \va^{'VLA})$. Moreover, we define the residual action such that the policy learns to correct the GROD VLA with the residual orchestrated actions. We train the policy with advantage weighted BC (AWBC,~\cite{peng2019advantage}) and then transition to the online RL phase. We evaluate this approach on the bowl and glass on rack environment and  report the evaluated performance (without VLM) of the trained policy in \cref{fig:AWBCLoss Distillation}. Here we observe that while the residual policy improves with the orchestrated data, when transitioning to the online RL phase, the policy learns significantly more slowly than the pure online RL baseline. We believe this is due to the distribution shift between the online RL and the offline RL/VLM distillation phase. 

\paragraph{Online RL with VLM Orchestration}
\begin{figure}[!ht]
\centering
\widefig{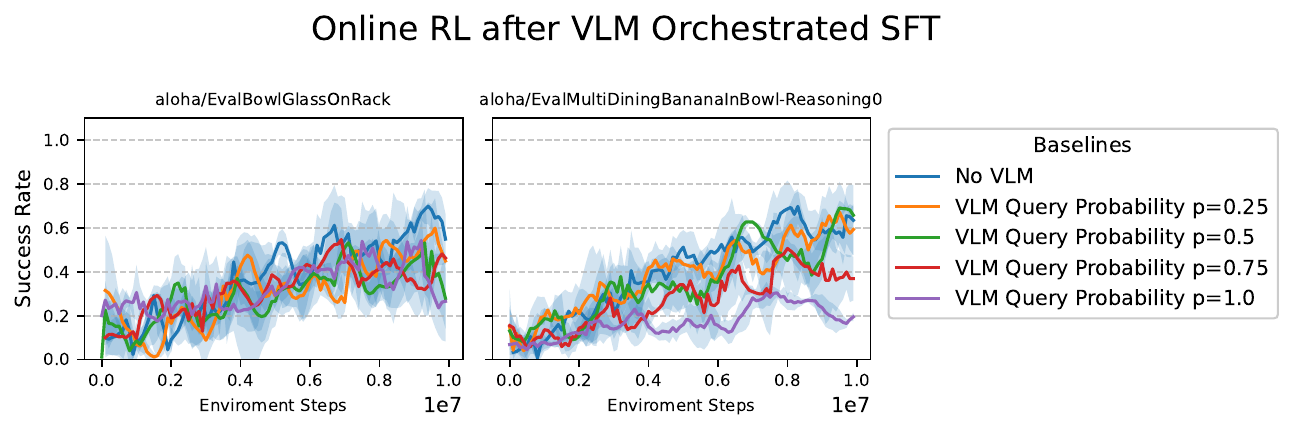}

\vspace{0.75em}

\widefig{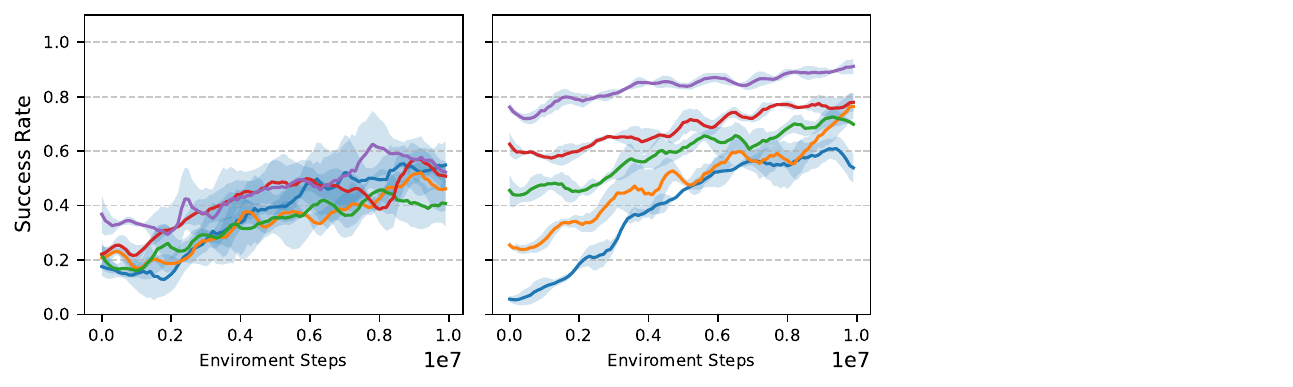}
    \caption{Performance of VLM to residual policy distillation where the VLM is only used with probability $p$ during the rollouts. 
    Top row: We observe that not using the VLM yields the best performance across the tasks during evaluation, despite the VLM orchestration leading to significantly more success in the initial phases of data collection (bottom row). We believe this is because the residual policy cannot benefit from the VLM orchestrated tuples. These results are obtained across three seeds and the mean performance with standard deviation is reported.}
    \label{fig:Offline RL VLM Turned off at random}
\end{figure}

In order to overcome the distributional shift, we train the residual policy with online RL and include the VLM orchestrator during the rollout phase. However, to overcome the distributional shift between train and evaluation phase (without the VLM), we only use the VLM for $p \times 100 \%$ of the episodes. Therefore, during training the VLM is only used for a fraction of the episodes collected. This allows us to collect on-policy data from the residual policy and also bridge the gap between the train and evaluation setting. We store the tuple $(\vx, \Delta \va, \vx', r)$, for RL training in the data buffer. Here $\vx = (\vs, \va^{VLA-VLM})$, and $\vx' = (\vs', \va^{'VLA - VLM})$. Note that when the VLM is not used $\va^{VLA-VLM} = \va^{VLA}$. The evaluation performance of the agent (without VLM) is reported in \cref{fig:Offline RL VLM Turned off at random} (top row). Here we observe that the method without the VLM performs the best across the tasks. We believe this is because the NO VLM baseline trains directly in the eval setting, i.e., learns to correct the base GROD model $\va^{VLA}$, where as $p$ transitions in the data buffer of the orchestrated method are for correcting the VLM orchestrated action $\va^{VLA-VLM}$. Despite the VLM orchestrated agent performing better during data collection (\cref{fig:Offline RL VLM Turned off at random} bottom row), the
residual policy does not benefit from the orchestrated tuples in the data and therefore performs worse than the non-orchestrated one during evaluation.  

\section{Implementation Details}\label{app:implementation}

\subsection{Task Suite}\label{app:tasks}
\Cref{tab:tasks} lists the 22 manipulation tasks used in our evaluation together with their natural-language goals. The task identifiers (T2--T23) match the per-task labels in \cref{fig:post_sft} and \cref{fig:results_post_rl} (T1 in those plots denotes the average over all tasks). The tasks span several categories: (i) \emph{dish-placement} tasks that require chaining multiple pick-and-place skills to place two items on a rack (e.g., a bowl and a glass), (ii) \emph{reasoning} tasks, in which the target object is not named explicitly but must be inferred from a semantic description (e.g., ``the item a monkey can eat''), (iii) \emph{spatial-understanding} tasks that require placing a tool in the correct (left or right) compartment of a caddy, and (iv) \emph{chained} tasks that require placing multiple tools into the caddy in sequence. The base VLA reliably solves the underlying atomic skills within its training distribution, but the compositional, reasoning, and spatial aspects of these tasks lie outside its training distribution, which is precisely where VLM orchestration helps.

\begin{table}[!ht]
\centering
\small
\begin{tabular}{@{}l p{0.37\textwidth} p{0.48\textwidth}@{}}
\hline
\textbf{ID} & \textbf{Task} & \textbf{Natural-language goal} \\
\hline
T2 & BowlGlassOnRack & put the bowl and glass on the rack \\
T3 & BananaInBowl-Reasoning0 & put the item that a monkey can eat into the bowl \\
T4 & MugOnPlate & put the mug on the plate \\
T5 & MugOnPlate-Reasoning0 & put the object you pour coffee in on the plate \\
T6 & MugOnPlate-Reasoning1 & put the object with a handle on top of the flat object \\
T7 & PenInContainer & put the pen into the white container \\
T8 & PenInContainer-Reasoning0 & put the object you use to write into the white container \\
T9 & PenInContainer-Reasoning1 & put the thinnest object into the white container \\
T10 & CanOpenerInCaddy-Left-Reasoning0 & place the can opener in the left compartment of the caddy \\
T11 & CanOpenerInCaddy-Right-Reasoning0 & place the can opener in the right compartment of the caddy \\
T12 & MagnifierCanOpenerInCaddy & put the magnifier and can opener in the caddy \\
T13 & MagnifierInCaddy-Left-Reasoning0 & place the magnifier in the left compartment of the caddy \\
T14 & MagnifierInCaddy-Right-Reasoning0 & place the magnifier in the right compartment of the caddy \\
T15 & ScissorsInCaddy-Left-Reasoning0 & place the scissors in the left compartment of the caddy \\
T16 & ScissorsInCaddy-Right-Reasoning0 & place the scissors in the right compartment of the caddy \\
T17 & ScissorsMagnifierInCaddy & put the scissors and magnifier in the caddy \\
T18 & ScissorsScrewdriverInCaddy & put the scissors and screwdriver in the caddy \\
T19 & ScrewdriverInCaddy-Left-Reasoning0 & place the screwdriver in the left compartment of the caddy \\
T20 & ScrewdriverInCaddy-Right-Reasoning0 & place the screwdriver in the right compartment of the caddy \\
T21 & ScrewdriverMagnifierInCaddy & put the screwdriver and magnifier in the caddy \\
T22 & PlateBowlOnRack & put the plate and bowl on the rack \\
T23 & PlateGlassOnRack & put the plate and glass on the rack \\
\hline
\end{tabular}
\caption{Manipulation tasks used in our evaluation and their natural-language goals. The task identifiers (T2--T23) correspond to the per-task labels in \cref{fig:post_sft} and \cref{fig:results_post_rl}; T1 in those plots denotes the average over all tasks. \emph{Reasoning} variants replace the explicit object name with a semantic description that the agent must ground to the correct object, while the left/right caddy tasks additionally require spatial understanding.}
\label{tab:tasks}
\end{table}

\subsection{VLM Orchestration Prompt}\label{app:prompt}
\Cref{fig:orchestration_prompt} shows the exact prompt template used by the VLM orchestrator (cf.\ \cref{fig:example_interaction}). At each step, the placeholders \texttt{[TASK]}, \texttt{\{base\_information\_description\}}, and \texttt{\{instruction\_guidelines\}} are filled in with the overall task goal, a description of the observation the VLM receives, and the constraints on the instructions it may emit, respectively. The latter two sub-templates are shown in \cref{fig:orchestration_subtemplates}.

\begin{figure}[!ht]
\begin{lstlisting}[style=promptstyle]
You are an expert robot programmer. Your task is to guide a robot to complete a task by providing continuous step-by-step natural language instructions. The robot uses a VLA policy that follows natural language.

The overall task the robot must achieve is: [TASK]

{base_information_description}

Based on the current state shown in the images, provide the *very next* instruction the robot should execute.

**Instruction Guidelines:**
{instruction_guidelines}

**Key Directives:**
1. Analyze the scene and the robot's state.
2. Determine the most immediate action to progress towards the task [TASK].
3. Provide *one* clear instruction per turn.
4. Explain your reasoning inside a <thinking> </thinking> block. Inside the <thinking> block, reason step-by-step:
      a. Review all camera viewpoints to understand the current scene.
      b. Identify all objects, their properties, and their spatial relationships relevant to the task: [TASK].
      c. Describe the robot's current state (e.g., hand positions, what it's holding).
      d. What is the overall task?
      e. Based on the goal and the current state, what is the single most logical and effective action the robot should take next to make progress?
      f. Formulate this action as a clear natural language instruction, following the Instruction Guidelines.
      g. Explain why this instruction is the appropriate next step.
5. Return your instruction inside a <score><instruct> block.

Example Output Format:
<thinking>
The task is to stack the red block on the blue block. The robot's right hand is free. It should first pick up the red block.
</thinking>
<score><instruct>pick up the red block with your right hand</instruct></score>

Now, analyze the current situation for the task: [TASK].
\end{lstlisting}
\caption{The main VLM orchestration prompt template used in \alg (cf.\ \cref{fig:example_interaction}). It queries the VLM for the next natural-language instruction given the current observations. The \texttt{\{base\_information\_description\}} and \texttt{\{instruction\_guidelines\}} sub-templates (\cref{fig:orchestration_subtemplates}) are substituted into this template at runtime.}
\label{fig:orchestration_prompt}
\end{figure}

\begin{figure}[!ht]
\begin{lstlisting}[style=promptstyle]
# {base_information_description}
You are provided images at different timesteps in the environment. For each timestep, you are given images from different camera viewpoints. You will also be given the history of your interactions with the user.
\end{lstlisting}

\begin{lstlisting}[style=promptstyle]
# {instruction_guidelines}
The instructions you provide must follow this structure:
1. Allowed commands are:
   a. `pick the [object description]`
   b. `put the [object description] in/inside the [container/location description]`
2. Describing Objects and Locations:
   a. Be specific. Always include color and the type of object or container (e.g., "the red block", "the blue bowl", "the wooden crate").
   b. Avoid generic terms like "thing" or "object" when a more specific noun is available.
   c. If there are multiple objects that fit the description, add details to disambiguate. Use positional language (e.g., "the leftmost green apple", "the block closest to the robot") or relational descriptions ("the cup to the right of the yellow banana").
   d. Do not refer to parts of the robot.
3. Provide the instruction in the <score><instruct> block.
   GOOD: <score><instruct>pick the red block</instruct></score>
4. One Action Per Instruction:
   a. Each instruction block must contain only a single `pick` or a single `put` command.
   b. Do not chain actions using "and", "then", or commas within a single instruction.
5. If the task is complete, only provide the <score>0</score>
\end{lstlisting}
\caption{The two sub-templates substituted into the main VLM orchestration prompt (\cref{fig:orchestration_prompt}): \texttt{\{base\_information\_description\}} describes the observations passed to the VLM, and \texttt{\{instruction\_guidelines\}} constrains the instructions the VLM may emit.}
\label{fig:orchestration_subtemplates}
\end{figure}

\end{document}